\documentclass[runningheads]{llncs}
\usepackage[T1]{fontenc}
\usepackage{graphicx}
\usepackage{float}

\begin{document}
\title{Investigating Model Compression for Neural Machine Translation in the Biomedical Domain}
\titlerunning{NMT in Biomedical Domain}

\author{
Maria Zafar\inst{1}\orcidID{0009-0008-9787-5634} \and
Souhail Bakkali\inst{2}\orcidID{0000-0001-9383-3842} \and
Rejwanul Haque\inst{1}\orcidID{0000-0003-1680-0099}
}

\authorrunning{M. Zafar et al.}

\institute{
\ South East Technological University, Carlow, Ireland \\
\url{https://www.setu.ie}, \url{https://www.compucore.ie} \\
\email{c00304029@setu.ie}, \email{rejwanul.haque@setu.ie}
\and
\ IRISA, France \\
\email{souhail.bakkali@irisa.fr}
}

\maketitle              
\begin{abstract}

Large-scale pretrained transformer models have achieved state-of-the-art performance across diverse machine translation tasks, including multilingual settings. Knowledge distillation has emerged as a sustainable approach for model compression, transferring knowledge from large teacher models to smaller, more efficient student models. Similarly, quantization—which reduces the numerical precision of model weights and activations (e.g., from 32-bit to 8-bit representations)—is widely used to accelerate inference, enabling models to run several times faster during deployment. However, both techniques face limitations when applied to specialized domain data, particularly under low-resource conditions. In knowledge distillation, the effectiveness of transfer is often constrained by the scarcity of domain-specific parallel data, while quantization can lead to performance degradation as bit precision decreases.
In this work, we investigate the combined application of knowledge distillation and quantization for French-to-English biomedical translation, a domain characterized by specialized terminology and limited parallel resources. We develop and compare multiple fine-tuning strategies to adapt compressed student models to this challenging setting. Our experiments demonstrate that a collaboratively distilled and quantized student model achieves a 69\% reduction in size, a 98.21\% increase in inference speed, and a 98.46\% reduction in CO$_2$ emissions compared to the original baseline—all without sacrificing translation quality. These results indicate that jointly optimized compression techniques can yield efficient, high-performance models suitable for translation service providers operating under resource constraints.

\end{abstract}
\section{Introduction}
Knowledge Distillation (KD) is a well-established technique for transferring knowledge from larger, more complex teacher models to smaller, more efficient student models \cite{hinton2015distillingknowledgeneuralnetwork,kim-rush-2016-sequence}. This approach involves training a student model to mimic the outputs of a teacher model, often using soft labels generated by the teacher on unlabeled data. Model quantization \cite{nagel2021whitepaperneuralnetwork} maps weights to a lower-bit data format, reduces memory footprint and accelerates model inference. 
The main goal of combining jointly KD and quantization is to create compact and deployable MT models that balance translation quality with inference speed
as this technique have proven effective in many MT translation tasks~\cite{moslem-2025-efficient}.
However, KD presents several challenges including (i) the severe lack of specialized parallel text data (e.g. biomedical) can prevent effective knowledge transfer, i.e., the student may overfit to the small data and fail to generalize the broad knowledge the teacher possesses; (ii) the shallow student model may not have the representational power for word sense disambiguation -it often falls back to the most frequent translation of a word-, leading to errors in disambiguation and to incorrect translation of polysemous terms. Moreover, quantization can significantly degrade translation quality for morphologically diverse languages and low-resource domains, especially under very low bit widths.

In this context, Translation Service Providers (TSPs) operating under constraint budget who offer online MT services face significant challenges since they require compact, fast and competitive MT models. Moreover, it is more challenging when the domain training data is not sufficiently available. 
In this paper we investigate applying KD with different fine-tuning strategies (e.g. QLoRA) \cite{dettmers2023qloraefficientfinetuningquantized} to develop compact, faster, deployable and competitive models for low-resource domain data translation. We employed quantization to the student models trained through KD in order to further reduce model size and inference cost. 
We carried out our experiments on French-to-English language pairs using ELRC-EMEA OPUS\footnote{OPUS: \url{https://opus.nlpl.eu/ELRC-EMEA/fr\&en/v1/ELRC-EMEA}} corpus. 
Our investigation showed that the quantized student model can perform comparably or even significantly outperform the baseline models. Note that the baseline models are those that are trained following the standard KD training setup. 

The rest of the paper is organised as follows. In Section \ref{sec:related} we describe the related work. Our datasets and details of experimental setups are explained in Section \ref{sec:setup}. In Section \ref{sec:experiment}, we discuss our results. Section \ref{sec:cnfw} concludes our work and discusses potential directions for future research. 

\section{Related Work}
\label{sec:related}

\cite{hinton2015distillingknowledgeneuralnetwork} introduced KD
for image classification task as a model compression technique to squeeze the large networks into smaller models which obtained the same performance as the large networks. Whereas \cite{kim-rush-2016-sequence} introduced KD techniques for neural MT (NMT).
Subsequently, a lot of research has been conducted on various applications of sequence-level KD for NMT
\cite{gu2018nonautoregressiveneuralmachinetranslation,kasai2020deep,gou2021knowledge,wang-etal-2021-selective,zhou2021understandingknowledgedistillationnonautoregressive}.
Most prominent works of \cite{kim-rush-2016-sequence} include sequence-level KD. This is an effective technique for training small student models based on pseudo target sequences generated by the large teacher models. The outcome of this technique effectively decreases both the size of the model and inference time, resulting in minimum loss in performance \cite{Bucila2006ModelC,kasai2020deep,sanh2020distilbertdistilledversionbert,gou2021knowledge,wang-etal-2021-selective}. 
KD generates forward-translated synthetic data (i.e. soft target labels) through the teacher model, and allows training of smaller models using synthetic data with a minimum loss in performance \cite{kim-etal-2019-research,yoo-etal-2021-gpt3mix-leveraging,schick2021exploitingclozequestionsshot,zhou2023scalablepromptgenerationsemisupervised}.
The approach is straight forward with two phases for the KD pipeline: the first phase generates distilled data using large language teacher models (i.e. LLMs) and the second phase uses the distilled data that is created during the first phase for training and fine-tuning the student models. 

Researchers have also investigated alternative prompting methods \cite{yao2023reactsynergizingreasoningacting,ho2023largelanguagemodelsreasoning} to improve sequence-level distillation by transferring reasoning and decision-making capabilities from teacher to student models. In most cases, sequence-level KD relied on correct or incorrect trajectories to fine-tune the student model \cite{li2023turningdustgolddistilling,chen2023mcckdmulticotconsistentknowledge,liu2024mindsmirrordistillingselfevaluation}. However, these studies mainly focused on large fine-tuning datasets and did not address scenarios with fixed teacher budgets or noisy teacher annotations in unlabeled data. To mitigate noise, \cite{yu2023training} rephrased noisy rationales to enrich fine-tuning data, but this approach did not consider selective annotation to reduce teacher cost.
\cite{britz-etal-2017-effective} proposed a setup where domain information was available during training but not at inference. They further developed a model by adding a domain classifier on top of the Neural Machine Translation (NMT) encoder, aiming to obtain a single system with performance close to multiple expert models. Although their method achieved comparable BLEU \cite{papineni-etal-2002-bleu} scores, it was less parameter-efficient than the target token model. BLEU evaluates the quality of MT model by comparing the translations of the model with the reference translations. Extending this idea, \cite{currey-etal-2020-distilling} introduced multi-domain KD, where multiple expert models were distilled into a single student. Their approach kept both architecture and capacity fixed, yielded substantial BLEU improvements over standard multi-domain models without increasing translation time or memory usage. Furthermore, their method was architecture-independent, allowing seamless integration with other multi-domain NMT frameworks while using fixed teacher depth and architecture.

Smaller NMT models offer advantages such as faster training, reduced computational demand, lower energy consumption, and suitability for deployment on resource-constrained devices, thereby contributing to lower $CO_2$ emissions \cite{kim-etal-2019-research}. However, these models often achieve lower translation quality compared to larger models, and the gap becomes more pronounced for low-resource languages. The main challenge in low-resource settings is the limited availability of high-quality parallel data. While large language models (LLMs) can generate soft target labels for knowledge distillation, their outputs for low-resource domains or languages are often noisy or of poor quality. This poses a problem for sequence-level knowledge distillation, as errors in the teacher’s outputs can be propagated, causing the student model to learn incorrect patterns from the distilled data.

Another model compression technique, quantization is also in focus by researchers. \cite{nagel2021whitepaperneuralnetwork} proposed a method for deploying complex AI models on resource-constrained devices by reducing their computational cost and energy consumption. The core idea involves converting the high-precision numbers (floating-point) typically used in neural networks into lower-precision integers, which significantly reduces memory footprint and computational requirements. 
\cite{gong-etal-2024-llmc} proposed LLMC, a toolkit designed for benchmarking and compressing LLMs. The core purpose of LLMC is to systematically explore the impact of quantization, a technique that reduces computational and memory demands of LLMs by converting them to lower-bit data formats, without significantly compromising accuracy. The toolkit offers extensive support for various algorithms, models, and hardware backends, enabling fair comparisons and detailed analyses of different quantization strategies. Key areas investigated include the influence of calibration data on performance, the effectiveness of various quantization algorithms (e.g. transformation, clipping, and reconstruction), and the trade-offs between integer and floating-point quantization. Ultimately, LLMC aims to provide researchers and users with practical guidance and insights to democratise LLM compression and facilitate their widespread application, even on more constrained hardware.
\cite{moslem-2025-efficient} investigated on methods for efficiently deploying large audio-language models for speech translation, addressing the significant computational demands of these powerful systems. Key approaches include iterative layer pruning, which systematically removes less important model layers, low-rank adaptation with 4-bit quantization (QLoRA) for more efficient fine-tuning, and KD, where a smaller student model learns from a larger teacher model. Their research demonstrates that combining these compression strategies can achieve up to a 50\% reduction in model parameters and storage, while retaining 97-100\% of the original translation quality for English-to-German and English-to-Chinese speech translation.
Most recently, \cite{zafar2025confidence} explored a novel approach to reduce the training costs and environmental impact of NMT for low-resource languages. They introduced a confidence-based KD method. This technique involves filtering out low-quality translations generated by a large teacher model NLLB-200-3.3B based on its confidence scores, thereby improving the efficiency of training smaller student models. Their experiments, focusing on Urdu-to-English translation, demonstrate that this strategy can significantly lower CO$_2$  emissions and training time without compromising translation quality. The research offers a practical and sustainable solution for translation service providers facing budget and resource constraints.

Our work focused on facilitating online MT translation companies with a reasonable solution for developing compact and fast MT models models without compromising the translation quality. In other words, we aimed at providing TSPs a cost-effective model compression solution with the blend of KD and quantization for MT, with applying different fine-tuning strategies. 
\section{Experiments}
\label{sec:setup}
This section discusses our configurations of the teacher and student MT models used in our experimental setup.
\subsection{Experimental Setups}
\label{sec:hypsetup}
We used MarianMT models \cite{junczys-dowmunt-etal-2018-marian} for our teacher--student training setups. 
We used the Helsinki-NLP/opus-mt-tc-big-fr-en\footnote{\url{https://huggingface.co/Helsinki-NLP/opus-mt-tc-big-fr-en}} checkpoint as our teacher model, which is referred as big transformer (BT). For building our student models, we used Helsinki-NLP/opus-mt-fr-en\footnote{\url{https://huggingface.co/Helsinki-NLP/opus-mt-fr-en}} checkpoint. From now on this model is referred as small transformer (ST).
Our training configuration is as follows: \textsf{\detokenize{batch size=8; eval batch size=8; learning rate=1e-5; epochs=20; 
save total limit=2; max sequence length=256; weight decay=applied}}
and the model was saved at every epoch \cite{zafar-etal-2024-setu-adapt}. To demonstrate the generability, we evaluate the performance of MarianMT models on the hold-out test data (cf. last row of Table \ref{tabstat}).

We quantized the models using CTranslate2\footnote{CTranslate2: \url{https://opennmt.net/CTranslate2}} for fast inference.
The conversion was performed using the \textsf{ct2-transformers-converter} utility, which transforms MarianMT models into a runtime optimised format compatible with CTranslate2. The quantization \textsf{float16} flag was applied to reduce numerical precision from FP32 to FP16 \cite{rajbhandari2020zeromemoryoptimizationstraining}. Naturally, this conversion significantly decreases memory consumption with notable improvements in inference speed.
We fine-tuned our MT models, and tested them under quantized and non‑quantized settings. 
Specifically, we used quantized teacher models for preparing distilled data to be used by the student model.
The intuition is to obtain a teacher model that can generate high-quality translations at faster inference speed.
To maintain the consistency we trained and tested all models with same configurations and on the computing system enabled with one type of GPU. We used A100 GPU throughout our experiments.


 
Note that we also fine-tuned the ST model on the training data (cf. Section \ref{sec:data}) before initiating the distillation process. 
We followed the standard student-teacher training setups in order to build our baseline student models. In other words, we treated the English translations of the French sentences as pseudo target labels for the distilled training data.

\subsection{Dataset}
\label{sec:data}

For our experiments we used French--English (Fr--En) parallel data from the biomedical domain, ELRC-EMEA OPUS.\footnote{OPUS: \url{https://opus.nlpl.eu/ELRC-EMEA/fr\&en/v1/ELRC-EMEA}} This is a medical domain data, covering a range of topics such as medicine, treatments.
The data statistics are shown in Table ~\ref{tabstat}. Note that we removed duplicates from the training data. We sampled a set of 1,000 and 2,000 source--target sentence-pairs from the same data source for the validation and test sets, respectively.

\begin{table} [H]
\small
  \centering
  \begin{tabular}{l r r r }
    \hline
    & Sentences & \multicolumn{2}{c}{Vocabulary} \\
   &  & French & English\\
    \hline
   Train & 759,861 &  85,627 & 69,036 \\
Valid & 1,000 & 4,454 & 3,963 \\
EMEA-Test & 2000& 4,557 & 4,052\\
    \hline \\
  \end{tabular}
 \caption{Data Statistics.}
  \label{tabstat}
\end{table}

\section{Results and Discussions}
\label{sec:experiment}
\subsection{Evaluation Metrics}
We used SacreBLEU \cite{post2018call} for measuring performance of our MT models. For reporting the results we used evaluation metrics such as: BLEU \cite{papineni-etal-2002-bleu}, chrF++ \cite{popovic-2015-chrF}, COMET \cite{rei-etal-2020-comet}, BERTScore (Precision, Recall, F1-score) \cite{zhang2020bertscoreevaluatingtextgeneration} and BLEURT \cite{sellam2020bleurtlearningrobustmetrics}. We measured inference speed through the time elapsed in seconds on the computing system enabled with A100 GPU.
We measured the energy consumption and carbon emissions~\cite{lacoste2019quantifying} during inference, for which we used carbon emission tracking tool, codecarbon\footnote{Code Carbon: \url{https://codecarbon.io}}
\begin{table}[H]
\small	
\caption{BT: Big Transformer. Performance of the BT (teacher) models on the EMEA test set. Inference Speed (IS): Total time required to translate test set data. Model Size (MS) on disk in MBs. SB: SacreBLEU. Q: Quantization. FT: Finetuning. P:Precision. R: Recall. F1:F1-score. BRT: BLEURT. Precision, Recall, and F1-Score is measured using BERTScore. Inference Speed is measured in seconds, CO$_2$ emission is measured in KGs during inference.}
\begin{center}
\begin{tabular}{|c|c|c|c|c|c|c|c|c|c|c|c|}
\hline
\textbf{Q} & \textbf{FT} & \textbf{SB} & \textbf{chrF++} & \textbf{COM} &
\multicolumn{3}{c|}{\textbf{BERTScore}} &
\textbf{BRT} & \textbf{MS} & \textbf{IS} & \textbf{CO$_2$} \\
\cline{6-8}
& & & & & \textbf{P} & \textbf{R} & \textbf{F1} & & & & \\
\hline
 no &no &51.27 &72.36 &87.90 &67.66&66.20&66.92&45.98&461&2983.88&0.0392\\\hline
yes &no &51.17 &72.94 &88.72 &77.46&76.77&77.08&46.58&442&41.05 &0.0006\\
\hline
no &yes &48.52 &70.24 &85.94&58.95&57.39&58.17&40.63&4.5&3777.21&0.0502\\
&  &(QLoRA) & &&&&&&&&\\
\hline
no &yes & 68.33 & 81.40 & 89.23 & 79.82 & 77.27  & 78.49 & 53.28 & 880 & 4147.61 & 0.0547 \\ 
\hline
yes &yes &\textbf{70.36} &\textbf{83.26} &\textbf{90.48}&\textbf{83.83}&\textbf{81.80}&\textbf{82.78}&\textbf{55.29}& 442&58.43&0.0008\\
\hline
\end{tabular}
\label{tab1}
\end{center}
\end{table}
\subsection{Teacher Models}
\label{sec:teacher}

As outlined in Section \ref{sec:hypsetup}, we used the MarianMT BT system \cite{junczys-dowmunt-etal-2018-marian} as the teacher for the translation task. Its performance on the test set is reported in the first row of Table \ref{tab1}.
We also evaluated the quantized version of this model, with the corresponding results shown in the second row of Table \ref{tab1}. 
We can see from the table that the quantized model outperformed the MarianMT BT system in terms of chrF++, COMET, precision, recall, F1-score while achieving comparable BLEU and BLEURT scores.

The third row of Table \ref{tab1} shows the performance of the MarianMT BT model after fine-tuning on the domain data (cf. training data; Table \ref{tabstat}) using QLoRA. 
Since the test set scores did not improve over the baseline models (first and second rows), we decided not to employ QLoRA-based fine-tuning for the teacher models in the subsequent experiments.

The fourth row of Table \ref{tab1} presents the performance of the MarianMT BT model after standard fine-tuning on the domain data. As can be seen from Table \ref{tab1}, the evaluation scores on the test set improved substantially compared to those in the first and second rows, with gains of 17.16 BLEU, 8.46 chrF++, 1.51 COMET, 12.16 precision, 11.07 recall, 11.57 F1-score and 7.3 BLEURT points over the baseline (i.e. first row of Table \ref{tabstat}). Like above, we also evaluated the quantized version of the fine-tuned BT model on the test set. The fifth row of Table \ref{tab1} represents the fine-tuned quantized BT model. When comparing the evaluation scores presented in fourth and fifth rows, we see additional gains of 2.03 BLEU, 1.86 chrF++, 1.25 COMET, 16.17 precision, 15.6 recall, 15.86 F1-score and 9.31 BLEURT points over those presented in the fourth row. 

We chose the best-performing BT model (i.e. corresponding to the fifth row of Table \ref{tab1}) as our teacher for the KD task. Quantization reduces the computational and memory footprint of the teacher during inference (see the last three columns of Table \ref{tab1}). This step not only facilitates faster model execution and lowers resource consumption, but also aligns with the broader aim of developing sustainable and efficient MT systems for an industrial translation pipeline.



\subsection{Student Models}
\label{sec:student}


\begin{table}[H]
\small	
\caption{Performance the Small Transformer models on EMEA test set. Small Transformer (ST). Inference Speed (IS): Total time required to translate test set data. Model Size (MS) on disk in MBs. SB: SacreBLEU. Q: Quantization. FT: Finetuning. P:Precision. R: Recall. F1: F1-score. BRT: BLEURT. Precision, Recall, and F1-Score is measured using BERTScore. Inference Speed is measured in seconds, CO$_2$ emission is measured in KGs during inference.}
\begin{center}

\begin{tabular}{|c|c|c|c|c|c|c|c|c|c|c|c|}
\hline
\textbf{Q} & \textbf{FT} & \textbf{SB} & \textbf{chrF++} & \textbf{COM} &
\multicolumn{3}{c|}{\textbf{BERTScore}} &
\textbf{BRT} & \textbf{MS} & \textbf{IS} & \textbf{CO$_2$} \\
\cline{6-8}
& & & & & \textbf{P} & \textbf{R} & \textbf{F1} & & & & \\
\hline
no  & no  & 52.91 & 73.59 & 88.50 & 76.86 & 75.55 & 76.17 & 48.24 & 298.00 & 1905.26 & 0.0250\\
\hline
yes & no  & 53.36 & 73.92 & 88.76  & 78.22 & 77.12 & 77.64 & 49.73 & 284.90 & 37.27   & 0.0004 \\
\hline
no  & yes & \textbf{65.09} & \textbf{79.96} & \textbf{81.04} & \textbf{81.73} & \textbf{79.04} & \textbf{80.34} & \textbf{55.73} & 284.90 & \textbf{1281.09} & \textbf{0.0167} \\
\hline
yes & yes & 66.79 & 81.44 & 89.66 & 84.00 & 82.17 & 83.05 & 57.99 & 142.90 & 52.56   & 0.0006\\
\hline
\end{tabular}
\label{tab2}
\end{center}
\end{table}

As mentioned in Section \ref{sec:hypsetup}, we used the MarianMT ST model \cite{junczys-dowmunt-etal-2018-marian} as the baseline system for KD in building the student model. We first evaluated the baseline model (MarianMT ST) on the test set. The first row of Table \ref{tab2} shows the performance of the baseline system (ST). We also evaluated the quantized version of the ST model shown in the second row. This model is comparable to the baseline as far as its performance on translation quality is concerned. The third row of Table \ref{tab2} represents the ST model that was fine-tuned on the training data. We see from the table that fine-tuning improved the ST model significantly, with an improvement of 13.88 BLEU, 7.85 chrF++, 1.16 COMET, 7.14 precision, 6.62 recall, 6.88 F1-score and 9.75 BLEURT points over the baseline ST model. The last row shows the scores that we obtained from ST model with fine-tuning on the training data. Note that this row refers to the quantized version of the ST model corresponding to the third row of Table \ref{tab2}. 
Quantization reduces the model to a lightweight, inference-only form and removes the ability to update its parameters. As a result, this model cannot be fine-tuned and therefore cannot be used as a student in KD. We report its performance for comparison only.
\begin{table}[H]
\small	
\caption{Performance of the Finetuned Student models on EMEA test set. DD: Distilled data, DDOP: Distilled Data with Original Parallel Data. ST: Small Transformer. Inference Speed (IS): Total time required to translate test set data. Model Size (MS) on disk in MBs. SB: SacreBLEU. Q: Quantization. FT: Finetuning. P:Precision. R: Recall. F1:F1-score. BRT: BLEURT. Precision, Recall, and F1-Score is measured using BERTScore. Inference Speed is measured in seconds, CO$_2$ emission is measured in KGs during inference.}
\begin{center}
\begin{tabular}{|c|c|c|c|c|c|c|c|c|c|c|c|}
\hline
\textbf{Q} & \textbf{FT} & \textbf{SB} & \textbf{chrF++} & \textbf{COM} &
\multicolumn{3}{c|}{\textbf{BERTScore}} &
\textbf{BRT} & \textbf{MS} & \textbf{IS} & \textbf{CO$_2$} \\
\cline{6-8}
& & & & & \textbf{P} & \textbf{R} & \textbf{F1} & & & & \\
\hline
no &DD &68.13 &81.42 &89.11&82.54&80.09&81.28&56.83 &284.90&1281.09&0.0167\\
\hline
yes &DD &\textbf{69.28} &\textbf{82.61} &\textbf{90.61}&\textbf{84.50}&\textbf{82.65}&\textbf{83.54}&\textbf{58.67} &\textbf{142.90}&\textbf{53.25}&\textbf{0.0006}\\
\hline
no & DDOP &66.69&80.49&89.58&80.02&77.13&78.52&53.94&284.90&1413.76&0.0182\\

\hline
\end{tabular}
\label{tab3}
\end{center}
\end{table}

The performance of the student models are reported in Table \ref{tab3}. 
The first row shows the results of our first student model. This model refers to the best-performing ST model (second last row of Table \ref{tab2}) was fine-tuned solely on the distilled data.
The second row of Table \ref{tab3} refers to the quantized version of this model. 
We can see from the table that quantization provided further improvements, adding up to 1.15 BLEU, 1.19 chrF++, 1.5 COMET, 1.96 precision, 2.56 recall, 2.26 F1-score and 1.84 BLEURT points over the non-quantized student model. 
The last row of the table shows the scores that we obtained by fine-tuning the best-performing ST model on the augmented data (original training data combined with the distilled data). Note that this is not a quantized model. We can see from the scores that this strategy does not work.

\begin{table}[H]
\small	
\caption{Best-performing teacher and student models. M:Models, OPD: Original Parallel Data, DD: Distilled data. BT: Big Transformer, ST: Small Transformer. IS: Inference Speed (IS): Total time required to translate test set data. Model Size (MS) on disk in MBs. SB: SacreBLEU. Q: Quantization. FT: Finetuning. P:Precision. R: Recall. F1:F1-score. BRT: BLEURT. Precision, Recall, and F1-Score is measured using BERTScore. Inference Speed is measured in seconds, CO$_2$ emission is measured in KGs during inference.}
\begin{center}
\begin{tabular}{|c|c|c|c|c|c|c|c|c|c|c|c|c|}
\hline
\textbf{M} & \textbf{Q} & \textbf{FT} & \textbf{SB} & \textbf{chrF++} & \textbf{COM} &
\multicolumn{3}{c|}{\textbf{BERTScore}} &
\textbf{BRT} & \textbf{MS} & \textbf{IS} & \textbf{CO$_2$} \\
\cline{7-9}
& & & & & &\textbf{P} & \textbf{R} & \textbf{F1} & & & & \\
\hline
BT& yes &OPD &\textbf{70.36} &\textbf{83.26} &\textbf{90.48}&\textbf{83.83}&\textbf{81.80}&\textbf{82.78}&\textbf{55.29}& 442&58.43&0.0008\\
\hline
ST& yes &DD &\textbf{69.28} &\textbf{82.61} &\textbf{90.61}&\textbf{84.50}&\textbf{82.65}&\textbf{83.54}&\textbf{58.67} &\textbf{142.90}&\textbf{53.25}&\textbf{0.0006}\\

\hline
\end{tabular}
\label{tab4}
\end{center}
\end{table}

The performance of the best-performing teacher and student models are reported in Table \ref{tab4}. When we compare the translation performance of best-performing teacher and student models (see Table \ref{tab4}), we can clearly see that the performance of the best-performing student model (e.g. last row of Table \ref{tab4}) is close to that of the teacher (e.g. first row of Table \ref{tab4}).

\subsection{Efficiency and Sustainability Analysis}
\label{sec:IS_CO}

We compared the teacher, and variants of the student models in terms of their inference speed, size (MB) and carbon emission during inference (test data). 
The BT and ST models differ significantly in terms of the costs when considering the size of the model, inference speed and carbon emissions generated during the inference. 
We refer the reader to the third-last, fourth-last and last columns of Table \ref{tab1} and \ref{tab4} for the differences in size of the models, the inference speed and CO$_2$ emissions of the best-performing student and teacher models. 


We can see that best-performing ST model (quantized version shown in second row of Table \ref{tab4}) reduced the size of model to 67.66\% and inference speed increased to 8.86\% over the best performing BT model presented in the first row of Table \ref{tab4}. We can see that the best-performing ST model increased inference speed to 98.21\% and decreased size of model to 69\% when compared to the baseline BT model shown in first row of Table \ref{tab1} but the difference in the quality can measured through the other metrics. 

The CO$_2$ emissions generated during inference of the BT and ST models were recorded. We can see that the best-performing ST model (shown in second row of Table \ref{tab4}) reduced the emissions of model to 25\% over best performing BT model shown in the first row of Table \ref{tab4}. We can see that the best-performing ST model can reduce CO$_2$ emissions by up to 98.46\% over baseline BT model shown in the first row of Table \ref{tab1} without any performance loss.

The online MT translation companies, which often operate under limited budget can benefit from our findings. KD when applied collaboratively with quantization with different fine-tuning strategies can result in competitive, compact, and fast MT models.

\section{Conclusion}
\label{sec:cnfw}

In this paper, we investigated the impact of fine-tuning and quantization together in a KD setup. We considered the context in which TSPs offering online MT services encounter significant challenges in obtaining compact, good quality models while translating specialised domain data. We conducted our experiments with different fine-tuning techniques such as QLoRA. We fine-tuned the teacher model and this ensured that our models were well adapted to the domain. For our distillation process we used quantized model for supervision. This sped up soft labels (translations) generation by the teacher model. Our results showed that the adaptation of the teacher to the domain data through quantization not only improved the translation quality or provided more reliable supervision for KD but facilitated faster model execution and lower resource consumption. This strategy could be beneficial for TSPs  who want to develop efficient, domain-adapted teacher model while reducing computational costs and maintaining translation quality.

When comparing the teacher and student models, we observed that the quantized and fine-tuned student model was able to achieve translation quality close to, and in some cases surpassing, that of the teacher. Despite being significantly smaller, the student maintained competitive performance while delivering much faster inference and lower emissions, confirming the effectiveness of our KD training setup. Our best performing student model obtained 98.21\% increase in inference speed and 98.46\% reduction in CO$_2$ emissions over teacher with insignificant drop in translation quality. This statistics show that our strategy aligns with the broader aim of developing sustainable and efficient pipelines for online MT service providers.
In future, we plan to extend our investigation to multilingual and low-resource language pairs to evaluate the generalizability of the proposed KD -- quantization approach. Additionally, integrating pruning and adaptive quantization techniques could further enhance model efficiency while maintaining translation quality.

%
%
%
\bibliographystyle{splncs04}
\bibliography{custom}





\end{document}